\documentclass[journal]{journal}

\usepackage{cite} 

\ifCLASSINFOpdf
\else
\fi 
\usepackage[colorlinks]{hyperref}
\usepackage{amsmath}
\usepackage[pdftex]{graphicx}
\usepackage{tabularx,ragged2e,booktabs,booktabs,caption}
\usepackage{threeparttable}
\usepackage{float}
 
\begin{document}
\title{Entropy Decision Fusion for Smartphone Sensor based Human Activity Recognition}
 
\author{{Olasimbo Ayodeji}~Arigbabu 
\thanks{e-mail: (oa.arigbabu@gmail.com).}}
\markboth{Journal of \LaTeX\ Class Files,~Vol.~6, No.~1, January~2007}%
{Shell \MakeLowercase{\textit{et al.}}: Bare Demo of IEEEtran.cls for Journals}

\maketitle
\thispagestyle{empty}

\begin{abstract}
Human activity recognition serves an important part in building continuous behavioral monitoring systems, which are deployable for visual surveillance, patient rehabilitation, gaming, and even personally inclined smart homes. This paper demonstrates our efforts to develop a collaborative decision fusion mechanism for integrating the predicted scores from multiple learning algorithms trained on smartphone sensor based human activity data. We present an approach for fusing convolutional neural network, recurrent convolutional network, and support vector machine by computing and fusing the relative weighted scores from each classifier based on Tsallis entropy to improve human activity recognition performance. To assess the suitability of this approach, experiments are conducted on two benchmark datasets, UCI-HAR and WISDM. The recognition results attained using the proposed approach are comparable to existing methods.
\end{abstract} 


\IEEEpeerreviewmaketitle

\section{Introduction} 
\IEEEPARstart{T}{he} increasing pervasiveness and wide acceptability of smart phones, coupled with immense amount of embedded sensors has open up an avenue for adopting mobile phones as a means of data acquisition. Transmitted signals from mobile sensors such as accelerometer and gyroscope can be captured for analyzing human motion based on acceleration of movement and angular rotational velocity.
 Aside data acquisition, recent advancement in edge intelligence (EdgeAI) \cite{Wang2019} has introduced another interesting perspective of developing self contained artificial intelligence device. EdgeAI offers on-demand prediction on smartphones using pre-trained models in real time with low latency, rather than depending on cloud deployment of trained models.  
 Such advancement provides compelling ecosystem to model human activity recognition (HAR) for fast and accurate personalization of activity pattern of an individual over time. This can then be incorporated into the developmental pipeline of systems used for video surveillance, patient rehabilitation, entertainment, and smart homes. 

Aside smartphones, there are various other means of capturing HAR data such as wearable sensors, ambient sensors, video sensors, and social network. \cite{Yeffet2009, Ikizler-Cinbis2010,Attal2015, Chen2015, Chen2020}. Like mobile, wearable sensors are usually placed at specific locations on the body (e.g sternum, lower back, and waist) to measure human activity information. Ambient sensors are usually installed in closed environment to monitor interactions between individuals and the environment. Video-based sensors are used to record daily activities of human subjects' appearance in video footage or surveillance cameras. Social network based activity recognition sifts through an abundance of users' online information from multiple social network platforms to understand users' intention, behavior, and activities. 
 
Human activities naturally possess inherent hierarchical structure, as an activity class can be divided into more fine grained smaller actions \cite{Ronao2016}, and activity data collected using sensors such as those in smartphones are quite complex to model. This is due to factors such as position of sensors, number of sensors, as well as complexity of activities, since different individuals may have slightly different styles of expressing a particular activity. Immense research effort has been devoted toward understanding and developing HAR by employing both handcrafted and deep learning based feature learning techniques to extract comprehensive information about different types of activities \cite{Palumbo2016}. To enable the usability of HAR in real life environment, a fundamental aspect which is very critical to its success is recognition accuracy. To improve the classification performance, different levels of information combination have been introduced which include sensor fusion, feature fusion, and classifier fusion \cite{Ehatisham2019, Nweke2019}. 

Multiple classifier fusion can help mitigate and compensate for the weakness of a single classifier, especially when the source of information is prone to a number of limiting factors that could negatively affect recognition accuracy.
However, an aspect of multiple classifier fusion that has not been well addressed is the generation of weights or significance factors assigned to different independent classifiers prior to score fusion. It is arguably an open area of research needing indepth investigation to improve HAR performance and boost its applicability in real life environment.

 Pushing toward using multiple learning algorithms for human activity recognition, this paper introduces a fusion strategy which integrates the predicted probabilities of convolutional neural network (CNN), recurent convoutional network (RCN), and support vector machine (SVM). 
Each learning algorithm is initially trained independently and the relative weighted scores from the learning methods are subsequently fused to produce a more robust and accurate activity recognition system. The relative weights in this case is based on computing self-information from each classifier using the Tsallis entropy \cite{Tsallis1988}, where the total Tsallis entropy from a classifier is influenced by the quality of information from each activity class. Therefore, ensuring that the final fusion weight for a classifier captures the underlying prediction quality of the classifier over the number of class labels in the model. 
 
The remainder of the paper is organized as follows: section II provides a review of related literature. Section III describes the adopted learning techniques based on CNN, RCN, and SVM. In section IV, we describe the fusion strategies and also present the proposed method for fusing the 3 learning methods. Experimental results are presented in section V, and we draw conclusions on this work in section VI. 

 
\section{Related Works}
This section describes the reported approaches in the literature for performing HAR from the perspective of feature representation and classification. It also points out the main contributions of this paper.
\subsection{Handcrafted features based methods}

It is commonplace in the literature to divide the pipeline of HAR into preprocessing, feature extraction, classification, which is also a standard in various computer vision and machine learning applications \cite{Arigbabu2015, Almaslukh2017}. Upon acquiring the data for HAR, a number of preprocessing operations are usually performed to minimize noise in the data. This includes the use of denoising filters, data segmentation and normalization techniques \cite{Wang2015}. From the perspective of feature extraction, algorithms such as principal component analysis (PCA), independent component analysis (ICA), and linear discriminant analysis (LDA) have popularly been used to extract statistical information from the raw data. Such is the approach presented by Fergani \cite{Fergani2015} where statistical feature extraction methods are used for training weighted SVM, with LDA-WSVM showing the best result. An overview of different feature extraction techniques was presented in \cite{Lara2012}, by grouping the techniques into structured and statistical. We have also witnessed direct application of machine learning classifiers to HAR by considering the input raw data as feature vectors. Some of the commonly used classifiers include decision tree, random forest, k-nearest neighbour (KNN), naive bayes, and SVM. Performance comparison of a number of both supervised and unsupervised classification techniques suggests that KNN and HMM are better suited for classifying human activities \cite{Attal2015}.

\subsection{Deep learning based methods}
Nowadays, the paradigm has shifted from deconstructing the learning process into sub-stages (preprocessing-feature extraction-classification) to using deep learning techniques which has the capacity to perform the aforementioned processes implicitly with very little requirement of human intervention in the learning pipeline. Deep neural network (DNN) \cite{LeCun2015} has attained ground breaking performance in other application areas such as image recognition, natural language processing (NLP), and object detection \cite{Liu2017}. The earliest attempt on HAR with DNN was presented in \cite{Plotz2011} using restricted Boltzmann machines (RBM). Recently, a lot of interesting research studies have been reported exploring various methods of learning useful information from data using DNN \cite{Ordonez2016, Murad2017, Nweke2018}. For instance, different architectures of deep, convolutional, and recurrent neural network (RNN) were assessed in \cite{Hammerla2016}. CNN with 1D filters and max pool layers was applied in \cite{Ronao2016} yielding a recognition result of 94.79\% and combination of CNN with temporal fast fourier transform (tFFT) produced a result of 95.75\%.  Comprehensive experimentation involving handcrafted features, codebook learning, multi-layer perceptron (MLP), CNN, LSTM, autoencoder, and CNN-LSTM on two public datasets was conducted in \cite{Li2018}. It was discovered that automatically learned features offer better performance than handcrafted ones especially using CNN-LSTM. To deploy DNN for HAR in real life applications, issues of speed has to be well addressed. This has been investigated in \cite{Inoue2018}, where deep RNN was trained on raw accelerometer data with high throughput of 1.347ms, which is 8.19 times faster than other similar methods. 

\subsection{Information fusion}
To improve the performance of HAR, researchers have opted for the option of integrating multiple sources of information. This has been approached from the perspective of fusing information from multiple sensors, where readings from accelerometer, gyroscope, and magnetometer are combined to compensate for the weaknesses of individual sensor \cite{Politi2014}. Such assessment has been conducted to understand the influence of positioning of on-body sensors at different body parts on HAR performance \cite{Shoaib2014}. There are studies also concentrating on fusion at feature level and classifier level \cite{Chen2020}. For instance,  CNN has been examined for implementing different early and late-fusion strategies such as integrating multiple CNNs trained independently on gyroscope, accelerometer, and pressure data at sensor, channel, and shared network parameter level \cite{Munzner2017}. Hierarchical decision fusion strategy was proposed in \cite{Banos2013} which typical combines multi-decision making classifiers by assessing the average accuracy of each classifier to obtain the weight for fusing the classifiers. Finally majority voting is used to obtain the final decision of each activity class \cite{Banos2013}. Using shannon entropy based weight generation, an approach which uses the classifier accuracy rates of multiple classifiers used under multiple sensors placed at different body to obtain classifier weights was proposed in \cite{Guo2018}. Handcrafted features were extracted from each sensor data, feature selection method based on LDA was adopted to minimize feature redundancy, and typical classifiers such as KNN, decision tree C4.5, naive bayes, SVM, and HMM were used to making predictions.
\subsection{Main contributions}
The contributions of this work can be summarized as follows:
\begin{itemize}
\item Explored deep learning and statistical learning based techniques to fuse their performance using the predicted class probabilities. Our method combines the usefulness of automatically learned features with handcrafted ones.
\item Proposed the use of generalized entropy based on Tsallis entropy to obtain classifier fusion weights which are directly influenced by relative self information of each classifier.
\end{itemize}

\section{Learning Methods}

This section describes the basic concepts behind the algorithms adopted in this work for training the activity data acquired using smartphone.

\subsection{ Convolutional Neural Network (CNN)}

CNN is a deep learning algorithm which combines feature learning with trainable classifier for learning from multiple array of data such as color image, video streams, or 3D volumetric data. \cite{LeCun1998, Krizhevsky2012}. The architecture of this type of network involves multiple layers of convolution, non-linear transformation, pooling operation, and a fully connected network at the tail end for prediction as depicted in Figure \ref{fig1a}. 
The convolutional layers in CNN perform convolution operation on the input data using a set of filter banks (kernels) with varying properties to generate some feature maps, which are use as input to the subsequent layers of convolution. This process is repeated until the entire convolutional layers have been exhausted. The units of the feature maps associated to an image patch are connected across different layers of the network with a set of weights \cite{LeCun2015}. Moreover, the feature maps are further approximated using a mapping (activation) function such as sigmoid, hyperbolic tangent, or rectified linear unit (ReLU) to ensure non-linearity. 

As an intermediate step between two convolutional layers, pooling operation is usually performed to generate features with strong semantic affinity. This strategically removes weaker values in the feature map as well as reducing the size of the feature map by replacing the values in a particular location with the statistical summarization of its neighboring features. This has also proven to help in eliminating the effect of variation caused by translation. Two main techniques have been suggested in the literature which are max pooling which replaces values of the feature map with the max value, and average pooling that simply computes the average of the feature map \cite{Maturana2015}. 
The final layers of CNN consist of fully connected network (FCN) and loss layer. FCN connects every single neuron in one layer to that of another layer, while the loss layer is used for making predictions. Softmax is usually used in the loss layer as it outputs the class probabilities for each class, particularly in a multi-class problem, between the range $[0, 1]$, and the sum of all the probabilities will be equal to one.

\begin{figure}[!h]
\centering
\includegraphics[height=6cm]{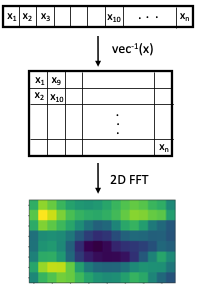}
\caption{Illustration of restructuring 1D data to 2D matrix. 2D FFT is then computed on the 2D matrix.}
\label{fig1}
\end{figure}

\begin{figure*}[!h]
\centering
\includegraphics[width=.8\textwidth,height=4cm]{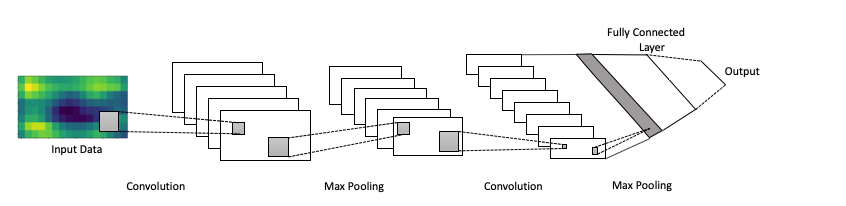}
\caption{Illustration of CNN with restructed data as explained in Section III and Figure \ref{fig1}. The convolutional layers of the network make use of 2D convolution, max pooling, and ReLU activation function, a fully connected softmax layer.}
\label{fig1a}
\end{figure*}

The type of data utilized in this paper is typically 1-dimensional on each axis as such 1D CNN is primarily used in this work. However, in a situation where we are interested in training the data with 2D CNN, we consider restructuring the 1D data to 2D. Suppose we can convert a 2D matrix $X  \in R^{m \times n} $ to 1D vector via vectorization as $vec(X) : R^{m \times n}  \rightarrow R^{mn}$, without loss of generality we can obtain the 2D matrix form of the vector by taking the inverse: $vec^{-1}(vec(X)) : R^{mn}  \rightarrow R^{m \times n}$, as shown in Figure \ref{fig1}. Afterward, we compute the 2D fast fourier transform (2D FFT) on the 2D matrix \cite{Gertner1988}, which then serves as input to 2D CNN. 
We understand that this may seem counter-intuitive since issues regarding coordinate directions while mapping from 1D to 2D should be put into consideration. However, from the empirical evaluation conducted, the recognition results attained using this approach is quite impressive. 

\subsection{Recurrent Convolutional Network (RCN)}
\begin{figure*}[h]
\centering
\includegraphics[width=0.95\textwidth,height=5.5cm]{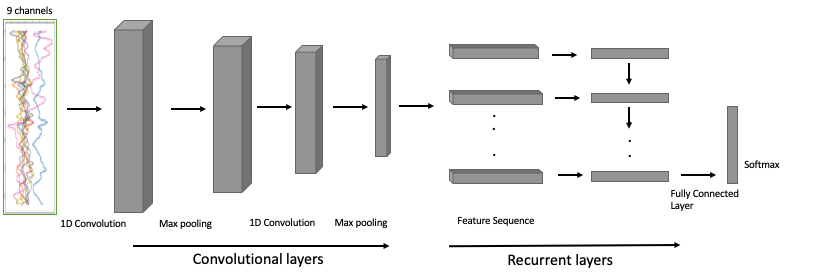}
\caption{Illustration of RCN with input data composing of 9 channel (axes). The convolutional layers of the network make use of 1D convolution, max pooling, and ReLU activation function. The resulting feature maps are fed to the recurrent layer (LSTM) and the final layer is softmax fully connected.}
\label{fig2}
\end{figure*}

RCN is basically the integration of recurrent neural network (RNN)\cite{Elman1990} and convolutional layers into a single learning framework. It has the advantages of both
convolutional and recurrent networks. RNN in its most traditional form attempts to construct a model with temporal dynamics flow by mapping sequential input data to a hidden state. The hidden states are then mapped to outputs which can be expressed with the following equations(\ref{eqn1}), given a sequence data $X$.

\begin{align}
\label{eqn1}
	h_{s} &= f(W_{xh}X_{s} + W_{hh}h_{s-1} + b_{s}) \\ \nonumber 
	z_{s} &= f(W_{hz}h_{s} + b_{z})
\end{align} 
where $f$ is a non-linear activation function computed element-wise, $h_s$ is the hidden state, and $z_s$ is the output at time $s$. One of the major challenges of RNN is the inability to remember interaction in long-term sequence due to the problem of exploding gradients \cite{Pascanu2013}. As a result, long-short term memory (LSTM) \cite{Hochreiter1997} networks have been introduced as a variant of RNN which incorporates memory units into the network. This effectively allows the network to determine the instances to forget previous hidden states or when to update hidden states when new data is fed into the network. In this work we construct a RCN with 2 convolutional layers of 1D filters and two layers of max pooling. The learned features are passed to LSTM and finally fully connected layer as shown in Figure \ref{fig2}.

\subsection{Support Vector Machine}
SVM is a structural risk minimization algorithm based on statistical learning theory \cite{vapnik1995}. The main concept is aimed at finding an optimal separating hyperplane that sufficiently separates the data. SVM has successfully been used in several machine learning classification problems such as image classification, face recognition, and object detection. We used the soft margin SVM which uses non-linear mapping functions to transform the data to high dimensional feature space and can separate the data by the introduction of some slack variables $\gamma_m$. 

\section{Fusion Strategy}

\subsection{Score Fusion}
To examine the impact of score fusion, three techniques are explored as follows. \\

\subsubsection*{\bf{Sum Rule}}

\begin{equation}
FS =  \sum^I_{i-1} S_i
\label{eqn4}
\end{equation}

\subsubsection*{\bf{Weighted Sum Rule}}
\begin{equation}
\label{eqn4} 
Weighted = \sum^I_{i-1} w_i S_i
\end{equation}
where, $S_i$ is predicted scores of a classifier and $w_i$ is a weight value assigned to the classifier based on recognition performance. \\

\subsubsection*{\bf{Entropy Weighted Score Fusion }}

we propose the concept of using self-informed classifier score significance for fusing multiple classifiers. 
Unlike weighted score fusion which generically assigns a weight value for each classifier, we instead compute the weight value dynamically by considering the amount of self-information each individual class label can contribute to the decision making. 

Assume that we have matrix of predicted class probabilities $S \in R^{d \times c} $ from a particular classifier (e.g SVM), $d$ is samples on the rows and $c$ represents the columnwise predicted probabilities of each class label with respect to each individual sample. Since, this is an obvious self contained indication of the classifier performance, we decide to compute the summation of entropies of the set of probabilities from the columns $c$. The proposed method is based on the Tsallis entropy, which is a generalization of the shannon entropy. As a measure of diversity of information, shannon entropy can be expressed as \cite{Shannon2001}:

\begin{equation}
H\left( c \right) = -\sum^n_{i=1} P(c_i)  log_2 P(c_i)
\label{shannon_eqn}
\end{equation}

where $P(c_i) $ represents the probability of possible outcomes of random variable $i$  in column $c$. 

With regard to the nature of distribution, shannon entropy makes implicit assumption about the tradeoff between contributions originating from the tails and the main mass of distribution \cite{Maszczyk2008}. It is however important to control such tradeoff explicitly to differentiate weak values coinciding with much stronger ones \cite{Maszczyk2008}. As a result, we propose using Tsallis entropy to obtain the weights for classifier score fusion. Due to its dependence on power of probabilities, Tsallis entropy allows us to control the contributions from the main mass and tail of the distribution with an \textit{entropic-index} parameter $\alpha$, which can be expressed as \cite{Tsallis1988}

\begin{equation}
H_\alpha \left( c \right) = \frac{1}{\alpha-1} \  \left(1- \sum_{i=1} P(c_i)^\alpha \right) ,  \ \ \  c_i > \tau
\label{renyi}
\end{equation}

where $H_\alpha(c)$ is a function for obtaining the Tsallis entropy of values $i = 1 \ldots n$ in column $c$. With the term $\alpha$, Tsallis entropy provides different level of concentration of information. $\alpha > 0$ will be more sensitive to values occurring more frequently, while $\alpha \leq 0$ will be sensitive to values occurring less often \cite{Wang2016}. The parameter $\tau$ we introduced in the equation is a small tunable parameter for selecting the predicted probabilities from the classifier, which are above a certain value. The main justification for parameter $\tau$ is to ensure that entropy is computed for values greater than $0$. This is because a classifier can return values within the range of $[0, 1]$ and entropy requires computation of logarithm, whereas the logarithm of $0$ is undefined.
Hence the entropy for a classifier is obtained using equation \ref{classifier}:

\begin{equation}
E_j =  \sum^c_{i=1} H_\alpha (c_i)
\label{classifier}
\end{equation}
$E_j$ is the summation of the entropies computed for each column in a classifier's predicted probability matrix.
In order to obtain the weight $\varphi_j$ of each classifier to perform score fusion, we use equation \ref{eqn5}:

\begin{equation}
\varphi_j = \frac{E_j}{E_{1} + \cdots + E_{j}}  
\label{eqn5}
\end{equation}
where, $\varphi_{j}$ is the relative weight for a classifier (such as SVM, RCN, and CNN), which is performed in turn for each classifier. To obtain the final fused score of the 3 classifiers, we simply apply the relative weight values to the predicted scores from the classifiers as expressed in equation \ref{eqn6}:

\begin{equation}
EWSF = \sum^J_{j-1} \varphi_j S_j
\label{eqn6}
\end{equation}
where $j$ is a classifier such as SVM, RCN, CNN.

\section{Experimental Results}
Experiments are conducted on two benchmark datasets, described as follows:
\subsection{Dataset}
\textsl{UCI-HAR}: the main dataset used in this paper is presented in \cite{Anguita2012}, which is collected by requesting 30 different subject to wear a smartphone (Samsung Galaxy S II) on their waists. Using the phone's accelerometer and gyroscope, tri-axial data of six different activities (walking, walking-upstairs, walking-downstairs, sitting, standing, laying) were collected. The data were sampled at a rate of 50 Hz, and separated into windows of 128 values, with $50\%$ overlap. In total there are 9 channels (axes) of gyroscope and accelerometer data, where each axis has a 128-real value vector activity depicting an activity. The 9 channels are: 
\begin{itemize}
\item[-]body accelerometer x-axis, y-axis, z-axis: 128 x 3 
\item[-]total accelerometer x-axis, y-axis, z-axis: 128 x 3 
\item[-]body gyroscope x-axis, y-axis, z-axis: 128 x 3
\end{itemize}

To conduct the experiments, we used the original split of the dataset composing of 7352 samples for the training and 2947 samples for testing. During training, we used random $10\%$ of the training data as validation set. Once training is completed, we then test the model with 2947 samples that were not used in training.

\textsl{WISDM}: the second dataset used in this work is WISDM \cite{Kwapisz2011} collected by recording 36 subjects performing 6 different activities such as walking, jogging, sitting, standing, climbing upstairs and downstairs. The dataset contains a total of 1,098,207 samples of one triaxial accelerometer sampled at a rate of 20 Hz. In addition, the authors of the dataset have included 43 extracted features based on each segment of 200 raw accelerometer readings, which are primarily based on average, standard deviation, average absolute
difference and time between peaks for each axis.

To conduct the experiments on WISDM we used 70\% of the dataset for training. Similarly, during training 10\% of the dataset are used as validation set. Once the model has been fully trained, the remaining 30\% of the dataset are used for testing.

\subsection{Settings}
For UCI-HAR data, in addition to the data restructuring performed in Section III, we used the original form of the first dataset \cite{Anguita2012} (which contains 9 channels of 128 accelerometer and 128 gyroscope values) and the preprocessed version of the data which consists of 561 values. Thus, the data configuration for the 3 learning algorithms are as followss: 
\begin{itemize}
\item[-] RCN: we used the raw input of 9 channels of accelerometer and gyriscope axes as input. The preceding layers of the network are composed of 2 layers 1D convolutional with rectified linear unit (ReLU) activation function, 2 layers of max pooling, and 1 layer of LSTM as illustrated in Table \ref{tab1}.

\begin{table}[!h]
\centering
\caption{RCN training paramters}
\begin{tabular}{l l}
\hline
Parameter & Value \\ \hline
Input data size & 128 \\ 
Input channels & 9 \\ 
Number of feature maps & 50-250 \\ 
Filter size & $[1\times 3]$-$[1\times 5]$\\ 
Pooling size & $1\times 3$ \\ 
Activation function & rectified linear unit (ReLU) \\ 
Learning rate & 0.01 \\  
Dropout & 0.2 \\ 
Batche size & 200 \\ 
Epochs & 200 \\  
LSTM cells size & 450 \\ \hline
\end{tabular}
\label{tab1}
\end{table}

\item[-] Scenario 2: involves using SVM for classification. The inputs to SVM are the preprocessed 561 features and the type of kernel used is radial basis function (RBF).

\item[-] Scenario 3: involves training 2D CNN on the restructured data explained in Section III. The 128 raw data from each axis is restructured into $16 \times 8$, resulting $16 \times 8 \times 9$ for the 9 axes as shown in Table \ref{cnn}.
\end{itemize} 

\begin{table}[!h]
\centering
\caption{CNN training paramters}
\begin{tabular}{l l l l}
\hline
Parameter & &  & Value \\ \hline
 & & 1D CNN (WISDM) & 2D CNN (UCI-HAR)\\ \hline
Input data size & & $ 43$ &  $16  \times 8 $  \\ 
Input channels & & 1 & 9 \\ 
Number of feature maps & & 10-80 & 20-100 \\ 
Filter size & & $[1\times 3]$ - $[1\times 5]$ &  $[2\times 2]$ - $[7\times 7]$ \\ 
Pooling size & & $1\times 2$ & $2\times 2$  \\ 
Activation function & & ReLU &  ReLU \\ 
Learning rate & & 0.01 & 0.001 \\  
Dropout & & 0.2 & 0.2\\ 
Batche size & & 200  & 300\\ 
Epochs & & 200 & 100 \\ \hline
\end{tabular}
\label{cnn}
\end{table}

For WISDM, we used the same parameters for training RCN, with the only variation in input data since the number of features is 43 with only one channel, SVM is trained using RBF kernel function, and 1D CNN is used as illustrated in Table \ref{cnn} 

\subsection{Results}
Examining the independent learning scenarios, we discovered that SVM with preprocessed data provided the best recognition performance of 96\% on UCI-HAR, RCN produced a result of 93.7\%, while 2D CNN with restructured data yielded the least performance of 91.9\%. This is quite surprising given that the raw tri-axial data typically possess features along 3 axes, which are not representative of coordinates in 2D matrix. Nevertheless, there seem to be some level of underlying structure in the 128 axial values which are somewhat transferable to 2D matrix. RCN on the other hand gracefully took advantage of the 1D form of the data, given that the convolutional layers are only 1-dimensional and LSTM itself is well suited to feature sequence.

\begin{table*}[!h]
\centering
\begin{threeparttable}
\caption{HAR results using different Learning Methods}
\begin{tabular}{l l l l l}
\hline
  Technique   & \textsl{UCI-HAR}&    &  \textsl{WISDM} &    \\ \hline
 & Acc (\%) & F1-score & Acc (\%) & F1-score \\ \hline
CNN & 91.9 &  91 & 81.7  & 81.5 \\  
RCN & 93.8 & 93.7 & 94  & 92.3  \\ 
SVM & 96 & 96  & 82 & 81 \\ 
Score Fusion & 94 & 94  &  86 & 84 \\
Weighted Score Fusion & 94.7 & 94.8  & 88.7 & 86.7 \\  
\bf{Proposed } & \bf{96.4} & \bf{96.3} & 89.5 & 89.4 \\ 
\bf{Proposed (RCN + SVM)} & \bf{97.4} & \bf{97.4} & 91.5 & 91 \\ \hline
\end{tabular}
\label{tab1a}
\begin{tablenotes}
\item $\tau = 0.1, \ \ \alpha = 2$  in all experiments
\end{tablenotes}
\end{threeparttable}
\end{table*}

\begin{figure}[!h] 
\centering
\includegraphics[width=.9\columnwidth]{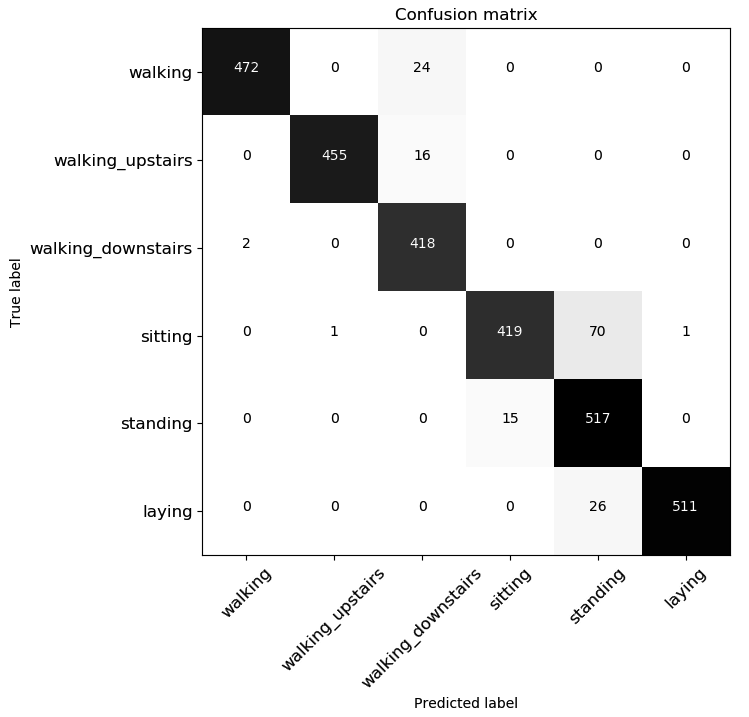}
\caption{HAR results using weighted score fusion.}
\label{fig3}
\end{figure}

\begin{figure}[!h] 
\centering
\includegraphics[width=.9\columnwidth]{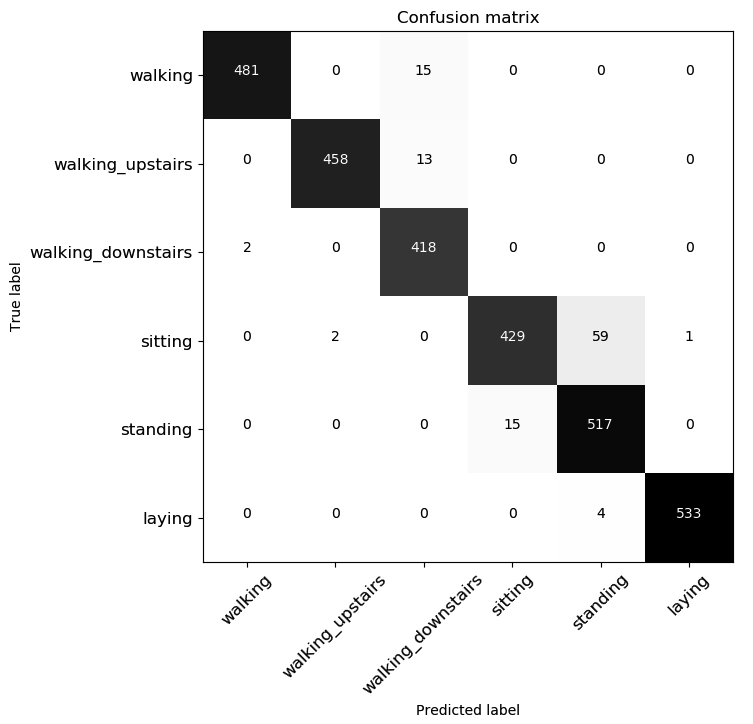}
\caption{HAR results using Entropy Weighted Score Fusion}
\label{fig4}
\end{figure}

In term of activity classes, RCN attain 100\% recognition on \textsl{walking-downstairs} activity and 84.11\% on \textsl{ssitting} activity which is its worst performance. CNN and SVM also showed their worst performance on \textsl{sitting} activity with recognition rates of 81.47\% and 89\%, while their best performance was attained on \textsl{laying} activity with recognition results of 94.97\% and 100\% respectively.  

With regard to decision fusion, the performance leveled out using ordinary score level fusion with a recognition rate of 94\%, weighted score fusion resulted in 94.8\%, the proposed entropy weighted score fusion with a performance of 96.4\% respectively. The confusion matrix of weighted score fusion and the propose method are depicted in Figure \ref{fig3} and \ref{fig4}. Though it can be argued that SVM is almost producing similar performance, however when we fused only RCN and SVM we attained a recognition result of 97.4\%. 

In the case of WISDM, RCN with a recognition result of 94\% significantly outperformed SVM and CNN whose results are 82\% and 81.7\% respectively. In terms of decision fusion, we noticed an increase in performance using weighted score fusion with recognition result of 88.7\%, while the proposed method attained a result of 91.5\%. 

The performance comparison of the proposed technique with state-of-the-art methods in the literature are presented in Table \ref{tab2a} and \ref{tab2b}

\begin{table}[!h]
\centering
\caption{Performance comparison on dataset \cite{Anguita2012}}
\begin{tabular}{l l}
\hline
Technique & Recognition Result (\%)\\ \hline
Random Forest \cite{Sousa2017} & 91 \\ 
SVM  \cite{Anguita2012} & 96 \\ 
Stacked Autoencoder \cite{Li2014} & 92.16 \\
CNN \cite{Ronao2016} & 94.79\\ 
tFFT + CNN \cite{Ronao2016} & 95.75 \\  
Proposed Method & 97.4 \\ \hline
\end{tabular}
\label{tab2a}
\end{table}

\begin{table}[!h]
\centering
\caption{Performance comparison on WISDM}
\begin{tabular}{l l}
\hline
Technique & Recognition Result (\%)\\ \hline
J48 \cite{Kwapisz2011} & 85.1 \\ 
Handcrafted + Dropout  \cite{Kolosnjaji2015} & 85.36 \\ 
Multilayer perceptron \cite{Kwapisz2011} & 91.7 \\  
Proposed Method& 91.5\\ \hline
\end{tabular}
\label{tab2b}
\end{table}

\section{Conclusion}
 
This paper has presented a new approach for performing decision fusion of classifier predicted scores of activity classes. The method involves computing self informed classifier score significance based on Tsallis entropy to obtain the weights for score fusion. We first examined the performance of independent learning algorithms on different structures of data captured using smartphone sensors from UCI-HAR and WISDM dataset. For this, we utilized RCN, CNN, and SVM, with SVM producing the best recognition result of 96\% on UCI-HAR and RCN 94\% on WISDM. We then assessed the proposed decision fusion method on the aformentioned benchmark datasets. From the experiments on UCI-HAR, we attained a recognition result of 97\% which is an improvement in comparison to independent classifiers, and other score fusion techniques. Moreover, the performance exceeded that of existing methods in the literature, while the performance on WISDM is quite competitive.

\end{document}